\documentclass[10pt,twocolumn,letterpaper]{article}

\usepackage{cvpr}              % To produce the CAMERA-READY version
\usepackage{multirow}
\usepackage{amsmath}
\usepackage{tabularx}
\usepackage{amssymb}
\usepackage{multirow}
\usepackage{bm}
\usepackage{booktabs}
\definecolor{cvprblue}{rgb}{0.21,0.49,0.74}
\usepackage[pagebackref,breaklinks,colorlinks,allcolors=cvprblue]{hyperref}

\title{LLV-FSR: Exploiting Large Language-Vision Prior for Face Super-resolution}

\author{Chenyang Wang, Wenjie An, Kui Jiang, Xianming Liu, Junjun Jiang}
\begin{document}
\maketitle
\begin{abstract}
Existing face super-resolution (FSR) methods have made significant advancements, but they primarily super-resolve face with limited visual information, original pixel-wise space in particular, commonly overlooking the pluralistic clues, like the higher-order depth and semantics, as well as non-visual inputs (text caption and description). Consequently, these methods struggle to produce a unified and meaningful representation from the input face. We suppose that introducing the language-vision pluralistic representation into unexplored potential embedding space could enhance FSR by encoding and exploiting the complementarity across language-vision prior. This motivates us to propose a new framework called LLV-FSR, which marries the power of large vision-language model and higher-order visual prior with the challenging task of FSR. Specifically, besides directly absorbing knowledge from original input, we introduce the pre-trained vision-language model to generate pluralistic priors, involving the image caption, descriptions, face semantic mask and depths. These priors are then employed to guide the more critical feature representation, facilitating realistic and high-quality face super-resolution. Experimental results demonstrate that our proposed framework significantly improves both the reconstruction quality and perceptual quality, surpassing the SOTA by 0.43dB in terms of PSNR on the MMCelebA-HQ dataset. 
\end{abstract} 
\section{Introduction}
Face super-resolution (FSR) is a technique that can recover the high-resolution (HR) face image from the low-resolution (LR) one. Due to the constraints of low-cost cameras and suboptimal imaging conditions, the captured face images are often of low quality, leading to poor visual effects and negatively impacting downstream tasks such as face recognition \cite{du2022elements}, attribute analysis \cite{zheng2020survey}, etc. FSR can improve image quality and boost the downstream tasks, 
%finding broad utility in the real-world applications. Therefore, FSR, standing as a critical technique in computer vision, 
which has gained more of spotlight in recent decades. %and been studied for many years.

Conventional model-based FSR methods rely on specific assumptions and prior and thus are less effective and practical on real complex scenes when the assumptions do not hold. Recent deep-learning based methods have shown considerable superiority over conventional algorithms in performance~\cite{jiang2021deep}. However, FSR is an ill-posed problem. Specifically, an LR face may correspond to multiple HR faces due to the difference in spatial dimensions, bringing great challenges to FSR task.

\begin{figure}[t]
    \centering
    \includegraphics[width=\linewidth]{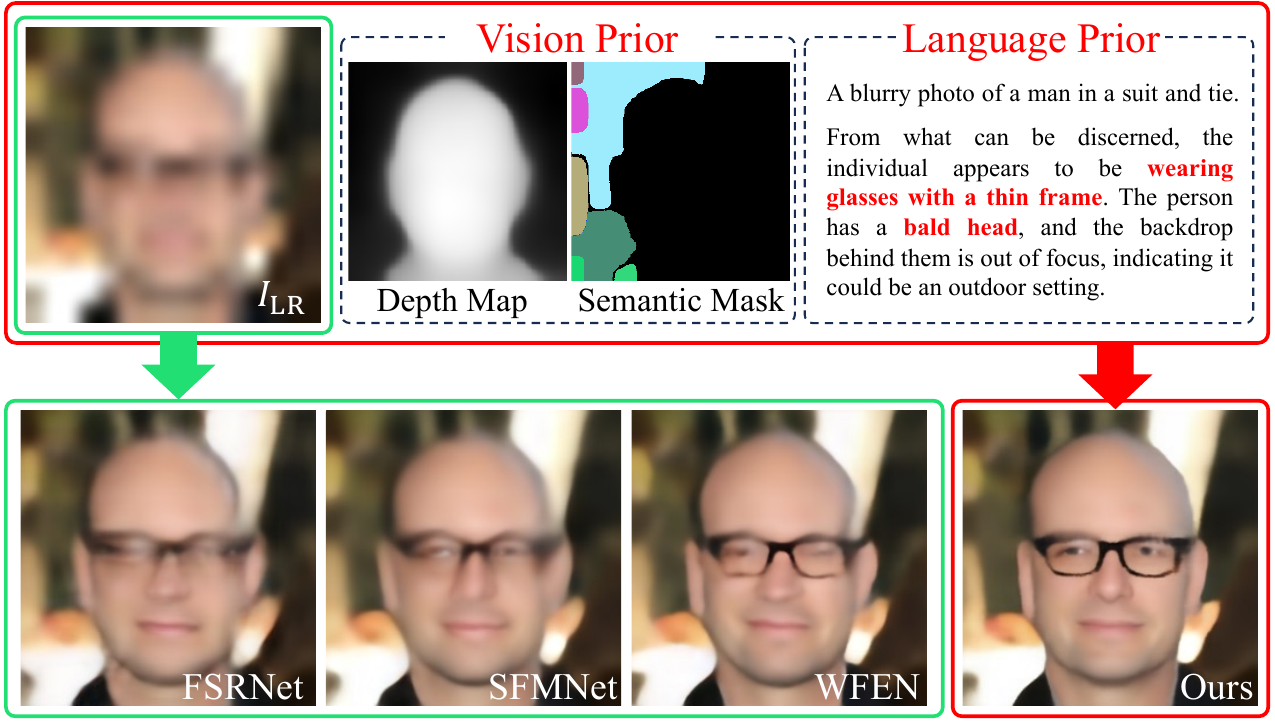}
    \caption{Comparison of FSR framework. The green part denotes existing FSR methods super-resoling face with original LR input while the red part presents our LLV-FSR generates high-quality output with language-vision prior.} %\vspace{-1em}
    \label{frame_com}
\end{figure}

Introducing additional prior is a common strategy for regularizing the solution space. 
For example, some efforts employ facial visual prior (\textit{i.e.}, facial parsing map, heatmap, and so on) to capture facial semantic structure information \cite{DIC,geometric,Chen_Lu_Yuan_Zhu_Li_Yuan_Deng_2024} for better reconstruction. 
Later on, generative prior-based FSR technologies exhibit remarkable capability 
to generate high-quality face images \cite{chan2022glean,DR2,glean++}, but not faithful to the ground truth. In addition, these methods have the following obvious drawbacks: i) overly reliance on specific individual visual prior, which is non-trivial to optimize the uncertainty of FSR; ii) predominantly centering around visual perception, while neglecting non-visual language textual information, which consequently results in incomplete scene representation and ultimately declines reconstruction performance of face images.

With the continuous advancement of large-scale models, some efforts have integrated language-vision prior into various computer vision tasks (\emph{e.g.}, video understanding \cite{ju2022prompting}), and achieved impressive and thrilling effects. Unlike the pixel-wise visual presentation, language (text) knowledge provides higher-level semantic understanding and abstraction which serve as a supplement to visual perception. This characteristic is valuable for ill-posed FSR problem and under explored in existing technologies. 

\textit{Based on the aforementioned analyses and observations, developing a unified framework to leverage more pluralistic language-vision prior  
for robust and photo-realistic FSR tasks is highly promising and worth exploring.}

Following this thread of thinking, we develop a novel framework LLV-FSR, which advocates for marrying language-vision prior for regularizing the FSR task, as depicted in Fig. \ref{frame_com}. Specifically, we generate the corresponding language-vision prior, involving image caption, description, facial semantic mask and depth map from the observed LR face with some pretrained large-scale models. Compared to limited visual knowledge (parsing map or depth) used in existing FSR methods, our language-vision prior balance visual semantics and advanced text abstraction expression, which are more comprehensive and in line with physical scene representation. Furthermore, we carefully design a language-vision prior fusion block to fully exploit the complementarity across language-vision features. Benefiting from the language-vision prior and elaborated fusion, it enables the network with powerful capacity to characterize the human face, thus enjoying the state-of-the-art FSR performance to generate visually pleasing FSR results. 

We highlight the contributions as follows:
\begin{itemize}
    \item We propose LLV-FSR, which makes the first attempt to marry the power of large language model and higher-order visual prior with challenging FSR task.

    \item  We carefully design an effect language-vision prior fusion block to utilize the complementary information contained in the language-vision representation, alleviating the ill-posedness of the FSR problem.

    \item Experimental results demonstrate that the proposed method achieves the state-of-the-art performance in terms of visual quality and quantitative metrics.
\end{itemize}

\section{Related Work}
\label{related_work}
\subsection{Face Super-resolution}
Face super-resolution (FSR) is a technique to recover high-resolution face image from the given low-resolution one. With the renaissance of deep learning in recent years, deep convolutional neural networks (CNNs) have pushed forward the frontier of FSR research. At early stage, researchers mainly design efficient architectures to transform the LR face into HR face directly. SCGAN \cite{hou2023semi} is designed recover real-world LR face images based on generative adversarial network. SFMNet \cite{SFMNet} leverages Fourier Transform to capture global facial structure and enhance the representability of the model. Considering the receptive field of CNN is local and transformer can capture global dependency, many transformer-based methods are developed. CTCNet \cite{ctcnet} builds a local-global feature cooperation module to capture local and global dependencies simultaneously, and then exploits the incorporation between them. 
Recently, diffusion-based FSR methods have gained more of the spotlight in research communities. IDM \cite{gao2023implicit} combines diffusion models with implicit representation to achieve continuous FSR while WaveFace \cite{waveface} proposes to recover different frequency components individually.

However, FSR is an ill-posed problem. It is very difficult to directly learn the mapping from LR to HR. Therefore, prior information is introduced to regularize the solution and assist FSR. Following this thread of thinking, prior-guided FSR methods are proposed. Different from natural image, face image has specific structural information as facial visual prior (\textit{i.e.}, facial landmarks, facial heatmaps, facial parsing maps, \textit{etc}.). For example, FSRNet \cite{FSRNet} enhances the LR face coarsely and then extract facial prior for further improving the quality of the intermediate results. DIC \cite{DIC} iteratively performs FSR and visual prior estimation for achieving collaboration between them. MFPSNet \cite{yu2023multiprior} adopts neural network to search efficient architectures for aggregating visual prior. Instead of utilizing 2D prior like previous methods, researchers exploit more informative 3D prior \cite{Chen_Lu_Yuan_Zhu_Li_Yuan_Deng_2024} to improve the quality of face images. 

Later on, researchers have discovered that pretrained face generative models can supply abundant facial information, which can be viewed as generative prior to enhance FSR \cite{gfpgan,glean++}. Initially, PULSE \cite{Pulse} directly optimizes the latent code to generative high-quality face whose downsampled version resembles the LR face. PaniniNet \cite{panini} extracts multi-scale features from LR face and feeds the features into the pretrained generative model to capture generative prior. SGPN\ cite{sgpn} explores the collaboration between geometric and generative prior for FSR. 
PD \cite{wang2023gan} proposes a pooling-based decomposition which can further improve the consistency of generative prior-based methods. Recently, DR2 \cite{DR2} employs a pre-trained denoising diffusion model to eliminate degradation artifacts before using enhancement models to refine high-frequency details. 
Although some progress has been made, they predominantly center around visual perception, while neglect non-visual language textual information, resulting in incomplete scene representation and limited FSR performance.

\begin{figure*}[t]
    \centering
    \includegraphics[width=0.99\linewidth]{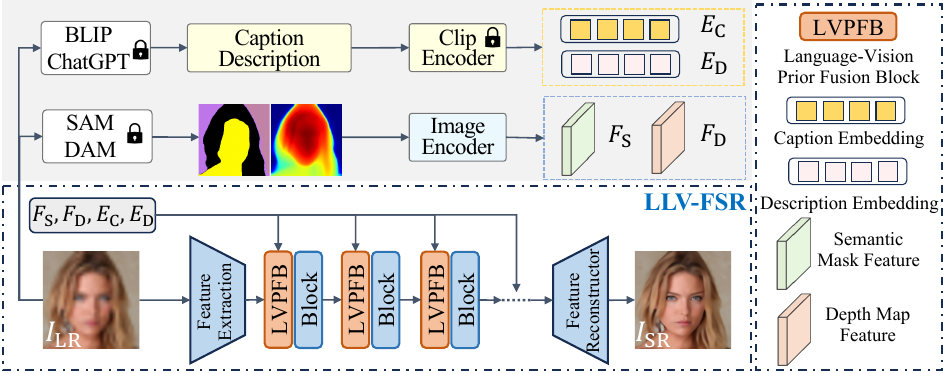}
    \caption{Overview of the proposed framework. Our method first inputs the LR face image into the pretrained large-scale model to extract language-vision prior and then exploit the prior information for improving face image quality.} \vspace{-1em}
    \label{VLPNet}
\end{figure*}

\subsection{Large-Scale Models}

Large-scale foundation models are typically trained on extensive datasets and generally exhibit superior performance in both computer vision and natural language processing. For example, Bert \cite{bert} is pre-trained on a vast amount of language data using mask-and-predict tasks to obtain language representation. SAM \cite{SAM} pre-trains a large segmentation model to generate semantic mask for input images while DAM \cite{DAM} exploits large-scale unlabeled data to train a depth estimation model for obtaining depth information.  
Recently, CLIP \cite{clip} strives to achieve the alignment between text and image. BLIP \cite{blip} introduces a two-stage approach to bridge the gap between image and text, and generates one sentence caption according to the input image. GRIT \cite{grit} leverages grid- and region-based visual features and establishes transformer-based network to generate better captions. ChatGPT \cite{achiam2023gpt} has the capability to generate detailed descriptions for any input image. These models can generate visual information (\textit{i.e.}, semantic mask or depth map) and language representation (\textit{i.e.}, captions or descriptions) for images, providing different representation of face images. In this paper, we investigate the potential of language-vision prior in FSR.

\section{Approach}
Face super-resolution (FSR) is committed to recover high-resolution (HR) face image from the given low-resolution (LR) one. To solve this ill-posed and challenging problem, existing methods either develop effective network architectures to learn the mapping from LR to HR, or introduce additional prior to constrain the solution space and assist face reconstruction. Although existing methods have made great progresses, they still have certain limitation. To be specific, they overly rely on specific individual visual prior, which is non-trivial to optimize the uncertainty of FSR. They predominantly center around visual perception, while neglecting language textual features, which consequently results in incomplete scene representation and declines reconstruction performance of face images. Recently, large-scale models have emerged and demonstrated exceptional capabilities for content generation. Different from visual prior utilized in existing FSR methods, language (text) knowledge depicts deeper understanding and higher-order abstraction, which can be viewed as language prior and serve as a supplement to visual perception. Therefore, we develop a unified framework to leverage the language-vision prior and explore their complementary information for recovering face images, dubbed as LLV-FSR. In this section, we would elaborate on the LLV-FSR.

\subsection{Overview}
Given an LR face image $I_{\text{LR}}$, we first feed it into the pre-trained large vision-language model to extract vision-language pluralistic representation. To be specific, we adopt the pre-trained BLIP2 \cite{blip}, ChatGPT, SAM \cite{SAM} and DAM \cite{DAM} to generate text caption, text description, semantic mask and depth map, respectively. The text caption can provide the overall face information and the text description depicts detailed facial content and features. While the semantic mask contains facial structure information and depth map depicts depth information. These prior complements each other and the combination of them presents face images more comprehensively, promoting face reconstruction. Thus, LLV-FSR incorporates these prior to assist FSR.
\begin{figure*}[t]
    \centering
    \includegraphics[width=\linewidth]{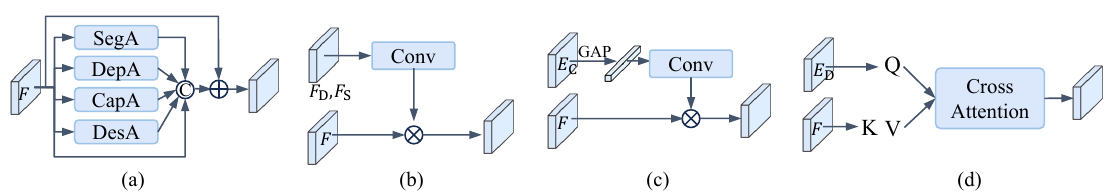}
    \caption{Framework of language-vision prior fusion block. (a): LVPFB; (b): SegA and DepA; (c): CapA; (d): DesA.} \vspace{-1em}
    \label{vlpfb}
\end{figure*}

The pipeline of LLV-FSR is illustrated in Fig. \ref{VLPNet}. With $I_{\text{LR}}$ as input, LLV-FSR first adopts a feature extraction layer (implemented by a convolutional layer) to extract visual features $F_{\text{0}}$, and then feeds the extracted features into the following L combinations of language-vision prior fusion blocks and basic blocks to fully exploit the potential of language-vision prior,
\begin{equation}
    F_{i}=f_{\text{Block}}^i(f_{\text{LVPFB}}^i(F_{i-1}, E_{\text{C}}, E_{\text{D}}, F_{\text{S}}, F_{\text{D}})),
\end{equation}

where $f_{\text{LVPFB}}^i$ and $f_{\text{Block}}^i$ denote the function of the $i$-th language-vision prior fusion block and basic block, and $F_{i}$ is the feature incorporated language-vision prior. Note that $E_{\text{C}}$ and $E_{\text{D}}$ are text embedding of caption and description extracted by the pre-trained text encoder of CLIP with fixed parameters while $F_{\text{S}}$ and $F_{\text{D}}$ are features of semantic mask and depth map encoded by image encoder implemented by a convolution layer. At last, we feed $F_{\text{L}}$ into a feature reconstructor implemented by a convolutional layer, generating the final super-resolved result $I_{\text{SR}}$. To constrain the LLV-FSR to recover satisfactory results, $\mathcal{L}_{\text{1}}$ loss is applied,
\begin{equation}
    \mathcal{L}_{\text{LLV-FSR}}=  \left\| I_{\text{SR}}-I_{\text{HR}} \right\|_{1},
\end{equation}
where $I_{\text{HR}}$ is the corresponding high-resolution face image.

\begin{figure*}[t]
    \centering
    \includegraphics[width=\linewidth]{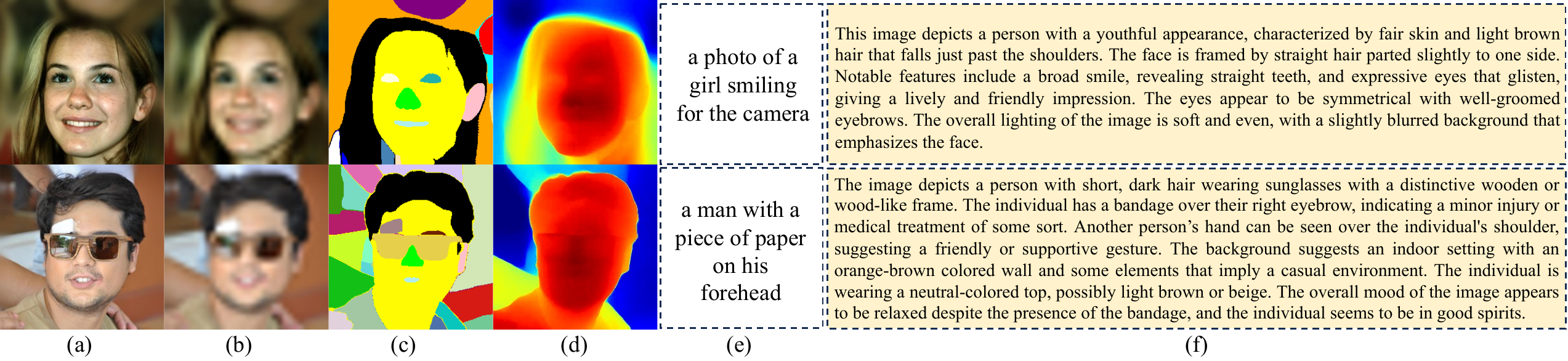}
    \caption{Language-vision prior visualization. (a): HR; (b): LR; (c): Semantic mask; (d): Depth; (e): Caption; (f): Description.} %\vspace{-1em}
    \label{mminfo}
\end{figure*}
\subsection{Language-Vision Prior Fusion Block}
Existing FSR methods center around visual perception by incorporating visual prior while neglecting language textual features, resulting in incomplete image representation and limited FSR performance. Recently large-scale models exhibit a remarkable capability of content generation and some studies have successfully leveraged language-vision prior into computer vision tasks. Complementary to visual prior, language textual knowledge depicts much higher-level understanding and abstraction, which is valuable for face reconstruction but underexplored in FSR. In this work, we construct a language-vision prior that combines both visual priors and higher-level language-textual priors to achieve a more complete and contextually rich representation of face images. This dual-prior approach not only strengthens the structural and visual fidelity of reconstructed faces but also introduces a layer of semantic depth that improves FSR performance by leveraging the interpretive power of textual information. To effectively integrate and capitalize on the complementary strengths of visual and language components, we have carefully designed a Language-Vision Prior Fusion Block (LVPFB), as illustrated in Fig. \ref{vlpfb}. This fusion block is specifically crafted to enable synergy between the visual and language-textual elements, maximizing the potential of the language-vision prior and enhancing FSR performance. Here we elaborate on LVPFB.

o effectively synergize visual and language components and exploit the potential and complementary information of language-vision prior, we carefully design a language-vision prior fusion block (LVPFB) as shown in Fig. \ref{vlpfb}. Here we elaborate on LVPFB.

Given the LR feature, and language-vision prior $E_{\text{C}}$, $E_{\text{D}}$, $F_{\text{S}}$, and $F_{\text{D}}$, LVPFB aims to integrate these elements and leverage their complementary information to enhance face super-resolution. Initially, LVPFB inputs LR feature into four parallel and distinct attention mechanisms (namely, SegA, DepA, CapA, and DesA) to interact with the four language-vision prior, respectively. Specifically, the depth map and semantic mask, which reflect facial structural and spatial pixel information, are processed by SegA and DepA. These mechanisms learn a facial structure-aware spatial attention by cascaded convolutional layers for enhancing LR feature. Instead of characterizing facial spatial information like visual prior, the text caption provides a global overview of the face image. Thus, global average pooling followed by convolution and sigmoid activation, is applied to the caption features to generate global caption attention, which is dubbed as CapA. Unlike the caption, text descriptions offer a more detailed and comprehensive insight into the face image. Accordingly, we use the text descriptions to generate queries Q, and the LR feature to generate keys K and values V. These are then fused using cross-attention, referred to as DesA. After integrating these four prior, we concatenate the four resultant features with LR features and introduce a skip connection to produce the final output. In LVPFB, we adopt different strategies to fuse different prior according their characteristics, achieving prior-aware and characteristics-adaptive fusion, then promoting FSR.

\subsection{Language-Vision Prior Generation}
Here, we introduce the detail in language-vision prior generation, including visual prior semantic mask and depth map, and language prior, text caption and description.

\textbf{Visual Prior: Semantic Mask and Depth Map}. 
A facial semantic mask captures high-level semantic information that delineates key structural regions of the face, such as the eyes, nose, mouth, and facial contours, allowing a detailed understanding of facial anatomy. Meanwhile, a depth map provides critical three-dimensional information about the spatial configuration of facial features, offering insight into facial depth variations and contributing to more realistic texture and shading in reconstructed images. Both elements are crucial for enhancing the representation of facial structure and facilitating FSR by introducing geometric and semantic cues.
In recent developments, large-scale foundational models, such as SAM \cite{SAM} and DAM \cite{DAM}, trained on extensive, diverse datasets, have demonstrated superior generalization capabilities and robustness against various types of image degradation \cite{xiao2023dive}. These models can extract semantic and structural details from degraded inputs with impressive accuracy. Therefore, we directly feed the LR face image into the pretrained SAM and DAM to obtain the facial semantic mask and depth map as visual priors.

  \begin{table*}[t] %\renewcommand{\arraystretch}{1.0}
   
  \centering
 \caption{ Comparisons with the state-of-the-art methods in terms of PSNR ($\uparrow$), SSIM ($\uparrow$), NIQE ($\downarrow$) and LPIPS ($\downarrow$) on the MMCelebA-HQ dataset. The best results are in bold and the second best results are underlined. }
  \begin{tabularx}{\linewidth} {X<{\centering}X<{\centering}X<{\centering}X<{\centering}X<{\centering}X<{\centering}X<{\centering} X<{\centering}X<{\centering}}
    \toprule

       \multirow{2}{*}{Methods}   & \multicolumn{4}{c}{$\times$8}&\multicolumn{4}{c}{$\times$16} \\ \cline{2-9}
        &  PSNR$\uparrow$&SSIM$\uparrow$  & NIQE$\downarrow$ &LPIPS$\downarrow$&PSNR$\uparrow$&SSIM$\uparrow$ & NIQE$\downarrow$&LPIPS$\downarrow$\\
    \hline
    Bicubic& 26.07&0.7156&12.335 & 0.2134&  22.88&0.6100& 15.700&0.2501\\
    SRCNN& 26.42&0.7240&12.016 &0.2030& 23.11&0.6079& 12.994& 0.2402\\
    FSRNet& 28.41&0.7874&9.741 &0.1317&24.85&0.6808&10.872 &0.1817\\

    DIC&  28.80&0.7982 & 9.692& 0.1251& 25.22&0.6930&\textbf{9.948 }& 0.1628\\

    SISN&28.73& 0.7959 & 9.460&0.1336 &25.09 &0.6878& 10.332&0.1901\\
 
    FaceFormer&28.88&0.8008&9.513 &0.1269 &25.47 &0.7021& 10.510& 0.1770\\

    SFMNet& \underline{29.12}&\underline{0.8104}&9.639&\underline{0.1140}&25.36 &0.6999 &10.417& 0.1743\\
    WFEN& 29.10&0.8089& \underline{9.450}&0.1182  &\underline{25.52}&\underline{0.7107}&10.406 & \underline{0.1610}\\

    Ours&\textbf{29.35}&\textbf{0.8140}& \textbf{9.419}&\textbf{0.1122}&\textbf{25.95}&\textbf{0.7228}&\underline{10.199}&\textbf{0.1501}\\

                 \bottomrule
\end{tabularx}   
   
  \label{comparison}
  \end{table*} 
\begin{figure*}[t]
	\centering
    \includegraphics[width=\linewidth]{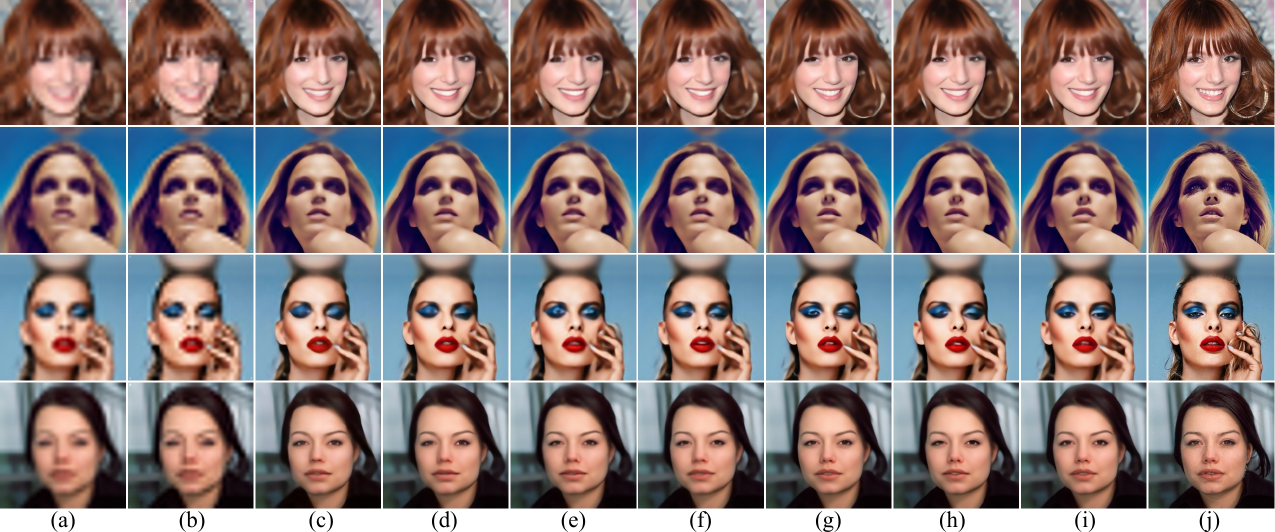}
 
	\caption{$\times$8 FSR results of state-of-the-art methods on MMCelebA-HQ dataset. (a): LR; (b): SRCNN; (c): FSRNet; (d): DIC; (e): SISN; (f): FaceFormer; (g): SFMNet; (h): WFEN; (i): LLV-FSR; (j): HR.}%\vspace{-1em}
	\label{fig:upscale_x8}
	
\end{figure*}

\begin{figure*}[t]
	\centering
    \includegraphics[width=\linewidth]{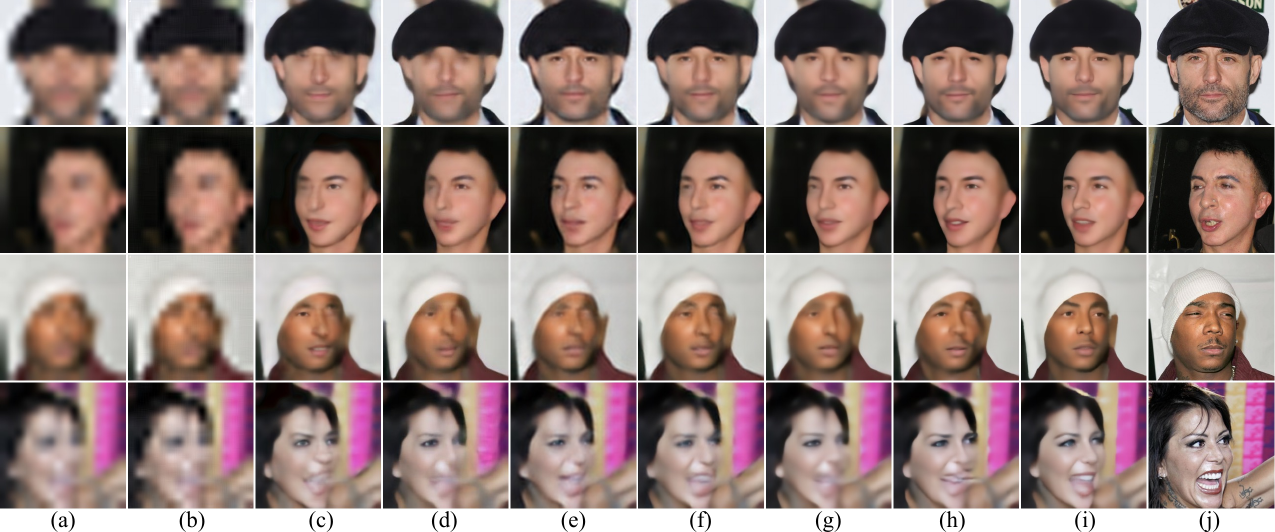}
 
	\caption{$\times$16 FSR results of state-of-the-art methods on MMCelebA-HQ dataset. (a): LR; (b): SRCNN; (c): FSRNet; (d): DIC; (e): SISN; (f): FaceFormer; (g): SFMNet; (h): WFEN; (i): LLV-FSR; (j): HR.}
	\label{fig:upscale_x16}
	
\end{figure*}
\textbf{Language Prior: Text Caption and Descriptions}. 
A text caption offers a global overview of a face image, encapsulating the image’s overall semantic essence and providing a concise yet meaningful summary of its primary features. This summary often highlights general aspects, such as gender, offering high-level information that complements visual representations. In contrast, a text description provides a more detailed and comprehensive account of the face, capturing finer nuances such as subtle facial expressions, specific skin texture, or any other particularities that may contribute to the face's individuality. This level of detail enriches the overall contextual understanding, providing insights beyond the visual elements alone. Fig. \ref{mminfo} illustrates some examples of language-vision prior knowledge. Unlike semantic masks and depth maps which depict structural information and depth information about the face, captions and descriptions deliver much higher-level semantic information and descriptions even contain more comprehensive and detailed textual information. These types of information—visual and textual—are distinct yet complementary, allowing for a richer and more comprehensive understanding of the face images. In this study, we feed LR images into BLIP2 to generate the corresponding image captions that summarize the face’s primary visual characteristics. For descriptions, the LR face image, along with a prompt that requests the model to provide useful visible features or observations, is inputted into ChatGPT-4, which then produces detailed text descriptions of the face images.

\section{Experiments}
\subsection{Datasets and Metrics}
We conduct experiments using the widely-recognized high-quality face dataset, MM-CelebA-HQ \cite{xia2021tedigan,xia2021towards}. Following the official setting, we use 24,000 face images for training and the remaining 6,000 face images for testing. To evaluate the model performance, four popular image quality evaluation metrics are chosen, including Peak Signal-to-Noise Ratio (PSNR), Structural SIMilarity (SSIM)~\cite{ssim}, Learned Perceptual Image Patch Similarity (LPIPS)~\cite{lpips}, Natural Image Quality Evaluator (NIQE) \cite{NIQE}.

\subsection{Implementation Details}
The original face images (256$\times$256) are directly used as ground truth. Then, we downsample the ground truth with bicubic interpolation into 32$\times$32 and 16$\times$16 to generate LR face images for $\times$8 and $\times$16, respectively. $L$ is set as 7. The basic block is implemented by two cascaded transformer blocks. The optimizer is Adam with $\beta_1$=0.9 and $\beta_2$=0.99, and $\epsilon$=1e-8. The learning rate is 2e-4. The model is trained for 30 epochs. The experiments are implemented on PyTorch with NVIDIA 4090 GPU with 24G memory.

\subsection{Comparison with the State-of-the-Arts}
We compare our method with the state-of-the-art FSR methods to verify the superiority of our LLV-FSR. In detail, we select several representative methods including the first deep learning-based super-resolution method SRCNN \cite{srcnn}, facial prior-guided methods FSRNet \cite{FSRNet} and DIC \cite{DIC}, convolutional neural network-based methods SISN \cite{SISN} and SFMNet \cite{SFMNet}, and recently proposed transformer-based methods FaceFormer \cite{faceformer} and WFEN \cite{wfen}. To be fair, all methods are trained and tested with the same dataset introduced above. The quantitative and the visual quality comparison are shown in Table \ref{comparison}, Fig. \ref{fig:upscale_x8} and Fig. \ref{fig:upscale_x16}.

\textbf{Quantitative Comparison}: Here we compare and analyze our methods from the prospective of the quantitative comparison and present the quantitative comparison results in Table \ref{comparison}. It can be observed that our LLV-FSR outperforms the existing methods. For example, on $\times$8 face super-resolution task, the PSNR of our LLV-FSR is 25.95 dB which is 0.43 dB higher than the second-best method WFEN. To be specific, FSRNet and DIC endeavor to estimate accurate facial visual prior and then incorporate facial prior. However, their performance is very limited. SISN develops inter split attention mechanism to capture facial information and it performs better than DIC and FSRNet. FaceFormer combines transformer and convolution neural network and its performance is better than SISN and DIC. WFEN and SFMNet adopt transformer and Fourier Transform to capture global receptive field and they further outperform SISN and FaceFormer. Although comparison methods perform well, they mainly focus on individual visual prior and pixel-wise representation while ignore the explore of more comprehensive language-vision prior. LLV-FSR incorporates language-vision prior into FSR task and characterize face images more comprehensively, leading to the state-of-the-art face reconstruction performance.

\textbf{Qualitative Comparison}: Fig. \ref{fig:upscale_x8} and Fig. \ref{fig:upscale_x16} illustrate the visual quality comparison of different methods for $\times$8 and $\times$16 FSR tasks. To facilitate a clear comparison, we present the LR and HR face images at the left and right column, respectively. As depicted in Fig. \ref{fig:upscale_x8}, SRCNN struggles to reconstruct key facial components. While FSRNet, DIC, and SISN can recover clear facial contours, they struggle with reconstructing critical facial details, such as the eyes and mouth, in the $\times$8 FSR task. However, their performance deteriorates significantly on the more challenging $\times$16 FSR task, yielding poor results. Similarly, in the $\times$16 FSR task, more advanced methods such as FaceFormer, SFMNet, and WFEN also show consistent failure in accurately restoring key facial components. These methods frequently generate distorted faces, with unnatural proportions or missing details that significantly impact visual quality. Although in certain instances (such as the first face in Fig. \ref{fig:upscale_x16}), some of these methods manage to restore certain facial features, the resulting images are often overly smooth and lack the intricate texture and sharpness. In contrast, our proposed method leverages a language-vision prior, which plays a critical role in enhancing the overall representation of the face. By incorporating both visual and textual information, our approach ensures that more comprehensive contextual and structural features are considered during the restoration process. This enables LLV-FSR to recover face images that not only have sharper and more accurate facial features but also exhibit a higher degree of realism and naturalness.

\begin{figure}[t]
	\centering

    \includegraphics[width=\linewidth]{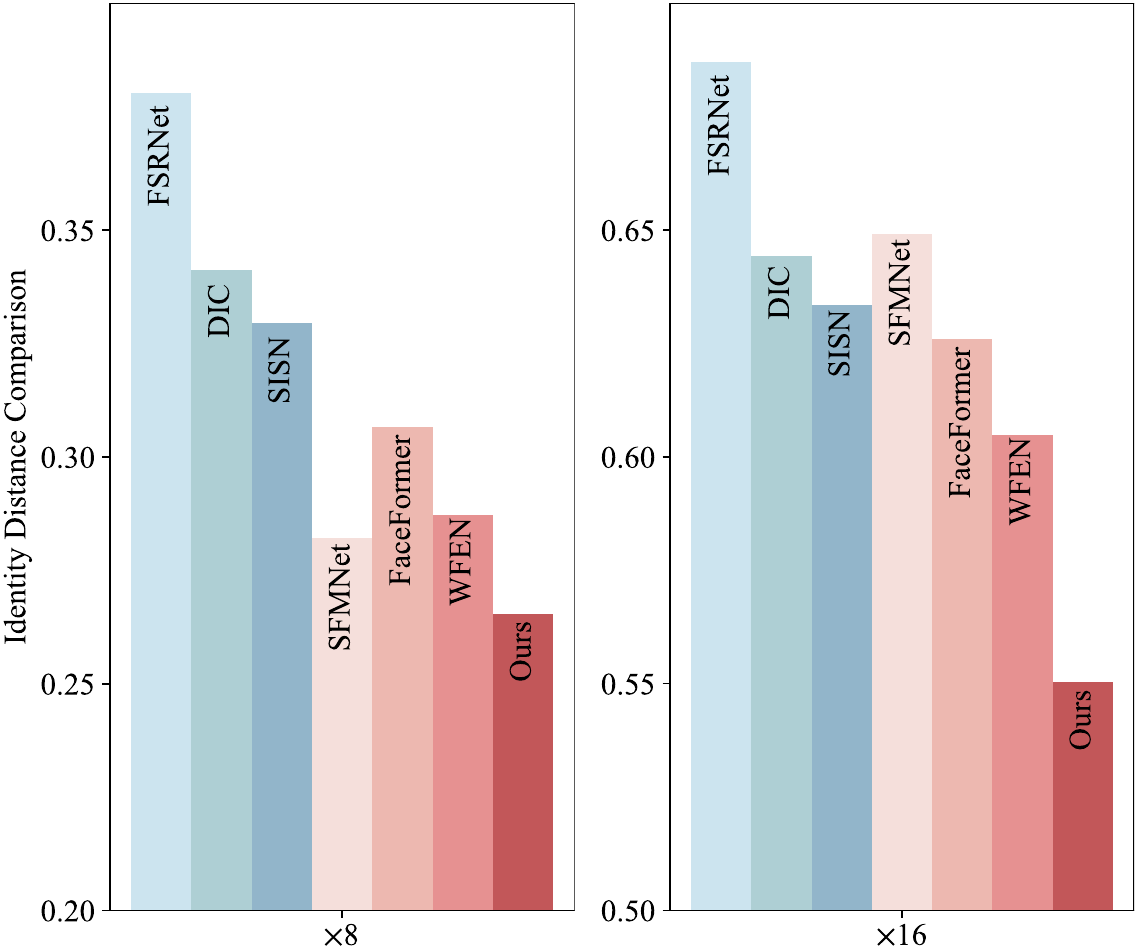}
 
	 \caption{Identity distance comparison.}
	\label{fig_idd}
	
\end{figure}

\textbf{Face Recognition Comparison}: In addition to improve the quality of LR face images, face super-resolution methods should also maintain facial identity information. That is to say, the super-resolved face images should share the same identity information with its corresponding HR one. Therefore, we further conduct comparison experiment in terms of identity distance. Specifically, we input the super-resolved face images recovered by different FSR methods and the corresponding HR ones into the pretrained face recognition model DeepFace \cite{deepface} to extract facial identity features. Then, the cosine distance between these features is calculated as a measure of identity distance. The comparison results are presented in Fig. \ref{fig_idd}. The identity distance of ours is lower than that of other methods, which demonstrates that LLV-FSR can better maintain the identity information of face image, then further improving face recognition task.

\subsection{Ablation Study}

\begin{table}[t]

    \centering
 \caption{Ablation study of the proposed LVPFB.}
     \begin{tabularx}{\linewidth}{m{1.4cm}<{\centering}X<{\centering}X<{\centering}X<{\centering}X<{\centering}}
    \toprule
      
           & PSNR$\uparrow$&SSIM$\uparrow$ & NIQE$\downarrow$&LPIPS$\downarrow$  \\\cline{2-4} 
      
         \hline

        Model 1& 29.17&0.8100&9.522 & 0.1143\\
        Model 2&29.24&0.8128&9.531 & 0.1129\\
        LLV-FSR&\textbf{29.35}&   \textbf{0.8140}& \textbf{9.419}&\textbf{0.1122}\\
         \bottomrule
    \end{tabularx}
   
    \label{ablation_VLPFB}
    
\end{table}

 \begin{figure}[t]
	\centering
    \includegraphics[width=\linewidth]{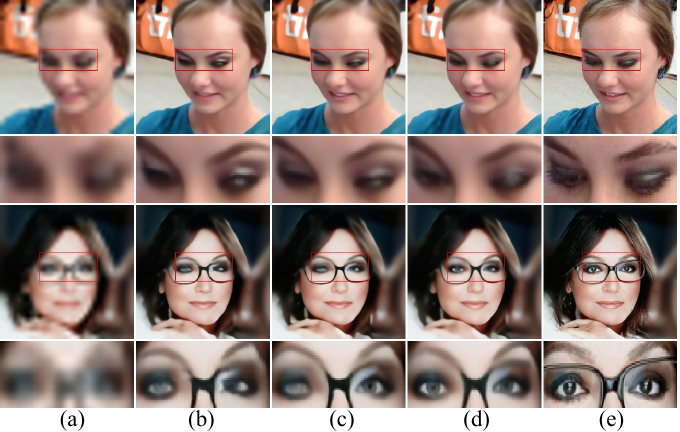}
	\caption{Ablation study. (a): LR; (b): Model 1; (c): Model 2; (d): LLV-FSR; (e): HR.}
	\label{fig_VLPFB}
	
\end{figure}

\textbf{Effectiveness of LVPFB}: Here we analyze the effectiveness of LVPFB. To be specific, we remove the LVPFB and the language-vision prior information, generating the baseline Model 1. Then we introduce the language-vision prior information and adopt concatenation followed by convolution layers to incorporate the language-vision prior, obtaining Model 2. Compared with Model 1, Model 2 achieves better performance, demonstrating that the introduction of language-vision prior information can boost face reconstruction. However, the performance improvement of Model 2 over Model 1 is limited. Then, we replace concatenation with our proposed LVPFB, dubbed as our LLV-FSR. From Table \ref{ablation_VLPFB}, LLV-FSR achieves the best quantitative performance and its improvement gain is obvious, demonstrating its effectiveness. We also illustrate the visual comparison of different models in Fig. \ref{fig_VLPFB}. In terms of visual quality comparison, LLV-FSR can recover more visually pleasing face images with more clear and accurate facial details, especially on facial eyes, than Model 1 and Model 2. To summary, the LVFPB can effectively fuse language-vision prior and enhance quantitative metrics and visual quality of faces.

\begin{figure}[t]
	\centering
    \includegraphics[width=\linewidth]{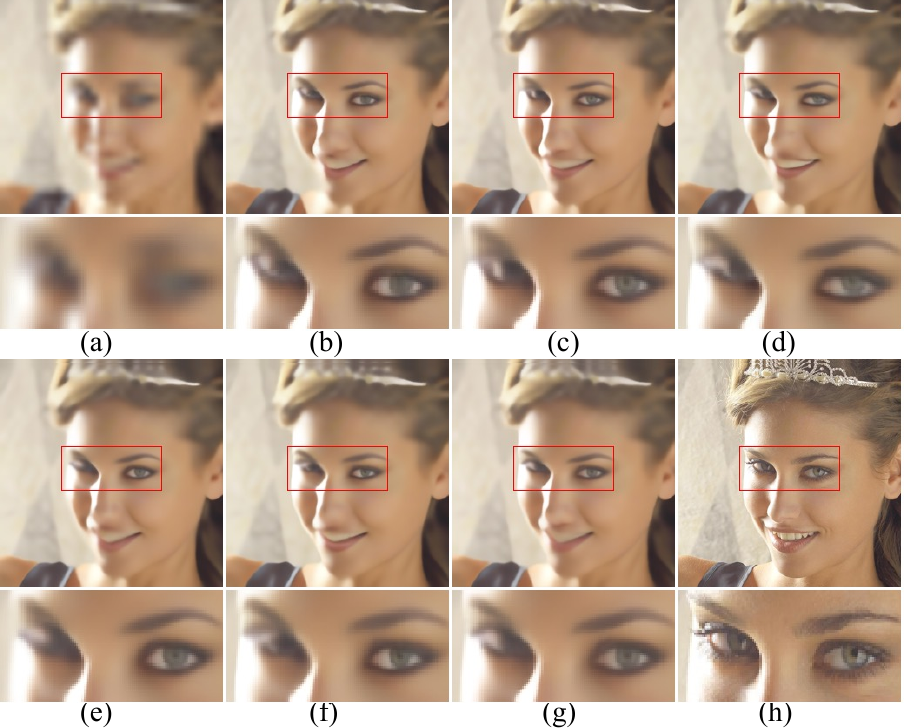}
	\caption{Ablation study. (a): LR; (b): LLV-FSR w/o prior; (c): LLV-FSR w/o $F_{\text{S}}$; (d): LLV-FSR w/o $F_{\text{D}}$; (e): LLV-FSR w/o $E_{\text{C}}$; (f): LLV-FSR w/o $E_{\text{D}}$; (g): LLV-FSR; (h): HR.}
	\label{fig_prior}
	
\end{figure}

\textbf{Effectiveness of Language-Vision Prior}: In this paper, we adopt four kinds of prior, including semantic mask, depth map, text caption and descriptions. Here we analyze the function of every prior. Fig. \ref{fig_prior} illustrates the faces recovered by models without differnt prior. Facial components of faces hallucinated by LLV-FSR without semantic mask $F_{\text{S}}$ or depth map $F_{\text{D}}$ is a little worse than LLV-FSR since they depict facial structure information which is important for FSR. The results of LLV-FSR without text caption seem similar to ones of LLV-FSR because the text caption reflects global information which is difficult to discern subjectively. LLV-FSR without $E_{\text{D}}$ struggles to reconstruct accurate details such as details of eyes, due to that all of semantic mask, depth and caption are unable to provide the detailed information description contains, and the absence of descriptions results in the loss of facial details. Overall, different prior contains different information and the combination of language prior and high-order visual semantic prior results in more comprehensive face representation, boosting FSR.

\subsection{Discussion and Limitations}

Although LLV-FSR can fully utilize language-vision prior and achieve the state-of-the-art performance, it is marred by a notable limitation: inference cost. In the inference stage, it needs to extract language-vision prior from the large-scale model, which requires some additional computation and time cost, limiting its application scenarios. In the future, we would further explore how to incorporate language-vision prior without additional computation cost.

\section{Conclusion}
In this paper, we propose a novel framework that effectively integrates language prior and visual semantics for face super-resolution task, dubbed as LLV-FSR. To be specific, besides directly absorbing knowledge from original input, we introduce the pre-trained large-scale model to generate pluralistic prior, involving caption, descriptions, semantic mask and depth map. These priors are then employed by our carefully designed language-vision prior fusion block to leverage their complementary information and guide more critical feature representation, facilitating realistic and high-quality face super-resolution. Experimental results demonstrate that the LLV-FSR can achieve the state-of-the-art performance and ablation study verifies the effectiveness of language-vision prior fusion block and the function of different prior information. 

{
    \small
    \bibliographystyle{ieeenat_fullname}
    \bibliography{main}
}

% WARNING: do not forget to delete the supplementary pages from your submission 

\end{document}